\documentclass[sigconf]{acmart}

\usepackage{graphicx}
\usepackage{amsmath}
\usepackage{epstopdf}
\usepackage{array}
\usepackage{multirow}
\usepackage{amsfonts}
\usepackage{booktabs}
\usepackage{subfigure}
\usepackage{bm}
\usepackage{algorithm}
\usepackage{algpseudocode}

\AtBeginDocument{%
  \providecommand\BibTeX{{%
    \normalfont B\kern-0.5em{\scshape i\kern-0.25em b}\kern-0.8em\TeX}}}

\begin{document}

\title{Reinforcement Learning for Weakly Supervised Temporal Grounding of Natural Language in Untrimmed Videos}

\author{Jie Wu}
\affiliation{%
	\institution{Sun Yat-sen University}
}
\email{wujie10558@gmail.com}

\author{Guanbin Li}
\authornote{Corresponding author is Guanbin Li. This work was supported by the State Key Development Program under Grant No. 2016YFB1001004, the Guangdong Basic and Applied Basic
Research Foundation under Grant No.2020B1515020048, National Natural Science Foundation of China under Grant No.61976250 and Grant No.61702565, the National High Level Talents Special
Support Plan (Ten Thousand Talents Program), the Fundamental
Research Funds for the Central Universities under
Grant No.18lgpy63, and was also sponsored by CCF-Tencent Open Research Fund.}   
\affiliation{\institution{Sun Yat-sen University}}
\email{liguanbin@mail.sysu.edu.cn}

\author{Xiaoguang Han}
\affiliation{\institution{Shenzhen Research Institute of Big Data, the Chinese University of Hong Kong}}
\email{hanxiaoguang@cuhk.edu.cn}

\author{Liang Lin}
\affiliation{\institution{Sun Yat-sen University}}
\affiliation{\institution{DarkMatter AI}}
\email{linliang@ieee.org}

\renewcommand{\shortauthors}{Jie Wu, et al.}

\begin{abstract}
Temporal grounding of natural language in untrimmed videos is a fundamental yet challenging multimedia task facilitating cross-media visual content retrieval.
We focus on the weakly supervised setting of this task that merely accesses to coarse video-level language description annotation without temporal boundary, which is more consistent with reality as such weak labels are more readily available in practice.
In this paper, we propose a \emph{Boundary Adaptive Refinement} (BAR) framework that resorts to reinforcement learning (RL) to guide the process of progressively refining the temporal boundary.
%BIR consists of three sub-networks, where the context-aware feature extractor encodes the environment state into multi-modal context concepts and the actor-critic action planner reasons a primitive action to adjust the boundary.
To the best of our knowledge, we offer the first attempt to extend RL to temporal localization task with weak supervision.
As it is non-trivial to obtain a straightforward reward function in the absence of pairwise granular boundary-query annotations, a cross-modal alignment evaluator is crafted to measure the alignment degree of segment-query pair to provide tailor-designed rewards.
This refinement scheme completely abandons traditional sliding window based solution pattern and contributes to acquiring more efficient, boundary-flexible and content-aware grounding results.
Extensive experiments on two public benchmarks Charades-STA and ActivityNet demonstrate that BAR outperforms the state-of-the-art weakly-supervised method and even beats some competitive fully-supervised ones.
\end{abstract}

%\begin{CCSXML}
%<ccs2012>
%  <concept_id>10010520.10010553.10010562</concept_id>
%  <concept_desc>Computer systems organization~Embedded systems</concept_desc>
%  <concept_significance>500</concept_significance>
%  <concept_id>10010520.10010575.10010755</concept_id>
%  <concept_desc>Computer systems organization~Redundancy</concept_desc>
%  <concept_significance>300</concept_significance>
%  <concept_id>10010520.10010553.10010554</concept_id>
%  <concept_desc>Computer systems organization~Robotics</concept_desc>
%  <concept_significance>100</concept_significance>
%  <concept_id>10003033.10003083.10003095</concept_id>
%  <concept_desc>Networks~Network reliability</concept_desc>
%  <concept_significance>100</concept_significance>
%</ccs2012>
%\end{CCSXML}
%
%\ccsdesc[500]{Computer systems organization~Embedded systems}



\maketitle

\section{Introduction}
Temporal grounding of natural language in untrimmed video is a newly-raised and crucial task due to its potential applications in the field of human-robot interaction and cross-media analysis. It aims to locate the temporal segment that is most relevant to the given sentence query in an untrimmed video.
Albeit with varying degrees of progress, most of its recent successes \cite{gao2017tall,liu2018attentive,ge2019mac,chen2019semantic,xu2019multilevel,yuan2018find,zhang2019man,wang2019language,he2019read,zhang2019exploiting,wujie2020AAAI} are involved in a fully supervised setting, i.e., mapping between video interval and the corresponding statement description are available in the training set.
It is still arduous to acquire such granular annotations that require a huge amount of manual effort, which becomes a critical bottleneck as this task is pushed toward a larger-scale and more complicated scenario. To alleviate such expensive and unwieldy annotations, \cite{mithun2019weakly} proposes to address this task in the weakly supervised setting that learns to infer language-related temporal range from video-level supervision.
This weakly supervised paradigm only has access to the video-level language description annotations without their corresponding temporal boundary specification.
This is an exceedingly favorable scheme since coarse video-level annotations are more readily available on the internet.
In our work, we focus on this weakly supervised paradigm.

\begin{figure*}[t]
	\centering
	\includegraphics[width=0.90\linewidth]{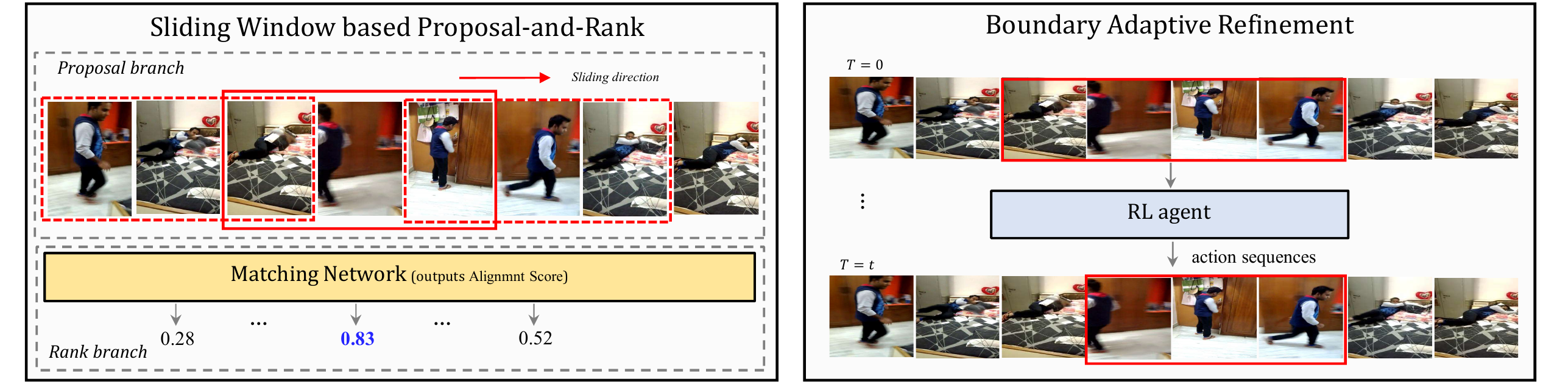}
	\caption{The diagrams of sliding window based proposal-and-rank pattern and the novel boundary adaptive refinement process. The input query in this example is ``person goes back to close the door.''
		Traditional pattern is constrained by fixed sliding window templates and has to process extensive candidate segments one by one to localize queries. However, the boundary adaptive refinement manages to flexibly adjust the boundary via a series of actions.
	}
	\label{fig:motivation}
\end{figure*}

Many approaches \cite{gao2017tall,liu2018attentive,ge2019mac,chen2019semantic,mithun2019weakly} employ a two-stage ``proposal-and-rank'' solution pattern to address the task of temporal grounding of natural language.
%They extract numerous sliding window based temporal segments and select the most appropriate one based on a pre-trained matching network.
However, these works are indulged in learning more robust cross-modal representations in the rank branch
without explicitly considering and modeling boundary-flexible and content-aware proposals.
As shown in the left half of Figure \ref{fig:motivation}, ``proposal-and-rank'' pattern is inherently restrictive as it relies heavily on pre-defined and inflexible sliding windows (e.g., 128 and 256 frames \cite{mithun2019weakly}), which results in lacking generalization for videos with considerable variance in length.
More rigorously, it raises two additional challenges when it is extended to the weakly supervised setting.
First,  offset regressive learning \cite{gao2017tall} for boundary adjustment becomes impractical in the absence of granular annotation.
Second, accessing video-query pair during training, the leading model \cite{mithun2019weakly} can merely learn cross-modal mappings from the inter-videos, while fails to take into account more subtle and fine-grained semantic concepts within the intra-video.
These suboptimal cross-modal mappings generally lead to less accurate boundary prediction.

To better cope with the above issues, as shown in the right half of Figure \ref{fig:motivation}, we formulate the task as a cross-modal matching guided heuristic process, a.k.a, \emph{ Boundary Adaptive Refinement (BAR)}. BAR resorts to a tailor-designed reinforcement learning paradigm to adaptively optimize the temporal boundary towards shrinking the cross-modal semantic gap. It is noted that reinforcement learning (RL) has been validated in various tasks of fully supervised video understanding, including video recognition \cite{yeung2016end} and video referring expression \cite{he2019read}. This work can be regarded as the first attempt to extend RL to weakly supervised temporal localization tasks. Due to the lack of matching supervision for specific video intervals corresponding to the statement,it is non-trivial to design an intensive learning state assessment and reward function which can effectively drive the model to achieve efficient temporal boundary optimization. Our proposed BAR framework hence includes a context-aware feature extractor for encoding the current and contextualized environment state, an adaptive action planner for decision adjustment of direction and interval range, and most typically a cross-modal alignment evaluator for providing an estimate of the alignment score between each segment-query pair in the absence of pairwise supervisory information. This alignment evaluator is crafted to assign a corresponding reward by comparing the alignment score of the consecutive segment-query pair under the guidance of both inter-video and intra-video ranking loss.
This modularized component design and heuristic adaptive temporal window adjustment strategy contributes to making the solution pattern more flexible and conforming to the human perception retrieval mechanism; Furthermore, it can be guided and pruned with goal-oriented rewards in a larger search space to extract more accurate temporal window positioning; Moreover, it also attempts to occupy as little time as possible to reach more impressive results.

The contributions of this work are summarized as follows:

$\bullet$ We design a Boundary Adaptive Refinement framework that resorts to reinforcement learning to address the task of weakly supervised temporal grounding of language in video. To the best of our knowledge, we are the first to employ RL to temporal localization
task with weak supervision.

$\bullet$ BAR abandons traditional sliding window based proposal-and-rank pattern and employs a novel boundary adaptive refinement process, which contributes to acquiring more efficient, boundary-flexible and content-aware grounding results.

$\bullet$ Experimental results on two benchmark datasets Charades-STA \cite{gao2017tall} and ActivityNet \cite{krishna2017dense} demonstrate that BAR outperforms the existing state-of-the-art weakly-supervised methods, and even beats some competitive fully-supervised ones.

%------------------------------------------------------------------------
\section{Related work}
\begin{figure*}[t]
    \centering
    \includegraphics[width=0.95\linewidth]{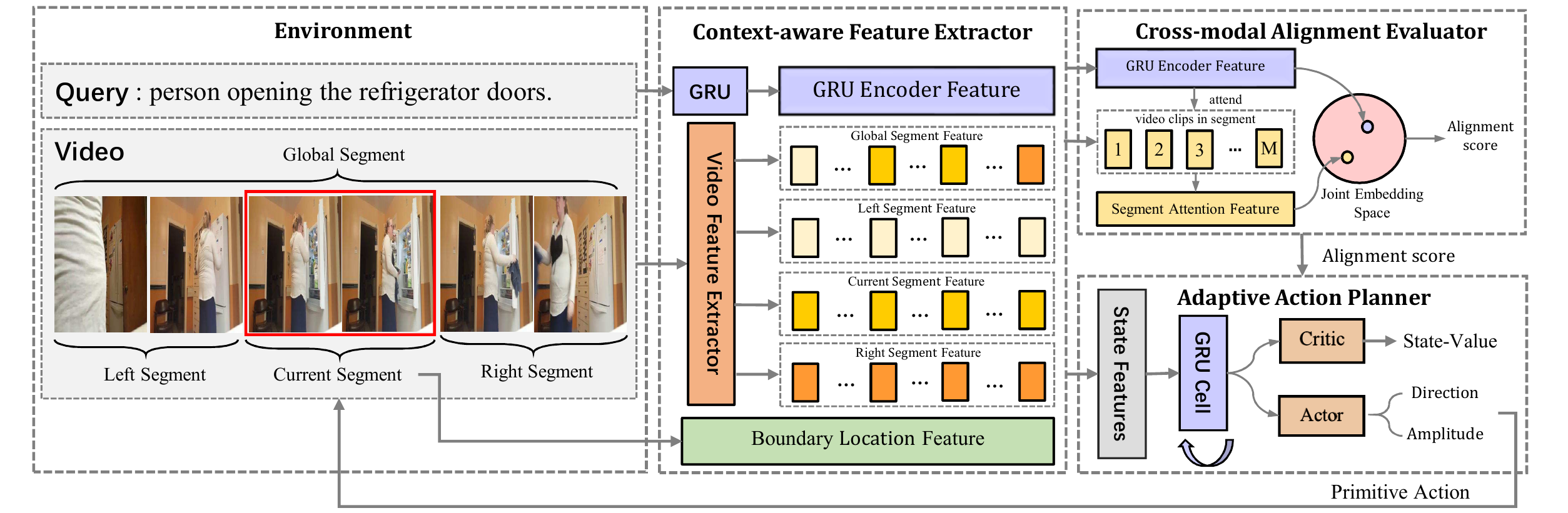}
    \vspace{-5pt}
    \caption{The overall architecture of the Boundary Adaptive Refinement (BAR) framework, which consists of a context-aware feature extractor, an adaptive action planner and a cross-modal alignment evaluator. }
    \label{fig:framework}
    \vspace{-5pt}
\end{figure*}
\noindent\textbf{Temporal Grounding of Natural Language in Video}.
Temporal grounding of natural language aims to determine the start and end time of a temporal segment in an untrimmed video that corresponds to a language query. It is a temporal extension of image referring expression comprehension\cite{yang2019cross,yang2019dynamic,yang2020graph}, and is also a challenging multimedia task which requires cross-modal fusion and fine-grained interactions between the verbal and visual modalities.
Many approaches \cite{gao2017tall,liu2018attentive,ge2019mac,chen2019semantic,mithun2019weakly} employ a two-stage ``proposal-and-rank'' manner, which first generates temporal proposals and then selects the one with the highest confidence score. However, these approaches rely on external sliding windows matching and ranking, leading to boundary-inflexible and time consuming.
To formulate a computationally efficient framework, Chen \emph{et al.} \cite{chen2018temporally} designed an end-to-end deep neural network that merely performs a single pass to obtain the grounding result.
Xu \emph{et al.} \cite{xu2019multilevel} proposed a multi-level model to integrate visual-query feature in the earlier stage and further introduced the caption generation as an auxiliary task.
%In this paper, we formulate the task as a sequential decision-making problem, which manages to adaptively adjust the initial temporal boundary and avoid inherent temporal templates.

\noindent\textbf{Weakly Supervised Learning}.
Weakly-supervised learning is a research setup that aims at optimizing a model without substantial manual labeled information.
Many computer vision and multi-modal tasks such as salient object detection \cite{li2018weakly}, captioning \cite{duan2018weakly}, language grounding \cite{paul2018w,mithun2019weakly}, referring expression grounding\cite{liu2019knowledge} have explored the weakly-supervised setup, since granular annotations are much more source-consuming compared to coarse annotations.
Wang \emph{et al.} \cite{wang2018collaborative} proposed a weakly supervised collaborative learning framework to resolve the task of weakly supervised object detection, which only requires image-level labels.
In the video domain, Duan \emph{et al.} \cite{duan2018weakly} formulated a new task: weakly supervised dense event captioning.
The goal of this task is to detect and describe all events of interest contained in a video without dense segment annotations for model training.
The work that closely related to ours is \cite{mithun2019weakly}. 
Mithun \emph{et al.} \cite{mithun2019weakly} 
designed Text-Guided Attention (TGA) mechanism
to leverage latent alignment between video frames and sentence descriptions to address the same task as us.

\noindent\textbf{Reinforcement Learning}.
Reinforcement learning (RL) is originated from the neuroscientific and psychological understandings of how humans learn to optimize
their behaviors in an environment. It can be mathematically formulated as a Markov Decision Process (MDP) in a sequential decision-making manner.
Recently, RL technique \cite{williams1992simple} has been utilized to imitate human's thinking pattern to address various tasks, which generally can be formulated as a MDP that executes a series of actions to accomplish the task-specific objective\cite{yu2019vision,ranzato2015sequence,he2019read,shi2019face,chen2018recurrent}.
Ranzato \emph{et al.} \cite{ranzato2015sequence} used the REINFORCE algorithm to train the captioning model in sequence level by optimizing the non-differentiable metric directly.
%Rennie \emph{et al.} \cite{rennie2016self} proposed a self-critical sequence training method to reduce the variance in RL algorithm.
Yeung \emph{et al.} \cite{yeung2016end} adopted REINFORCE algorithm to optimize an end-to-end approach for reasoning the temporal bounds of action and its category.
He \emph{et al.} \cite{he2019read} resorted to RL based method to address the fully-supervised version of the studied task, which utilized the temporal IoU as reward indicator.
Our work offers the first attempt to extend RL to accomplish the proposed task with weak supervision.
To estimate an accurate reward function in the absence of pairwise supervisory information, a cross-modal alignment evaluator is crafted to provide tailor-designed rewards.

\section{Methodology}
\subsection{Problem Formulation}
Following the common-used formulation~\cite{gao2017tall}, we represent a video $V$ by $N$ clips $\{V_1,V_2,...,V_N\}$, each clip corresponds to a small chunk of sequential frames. Taking $V$ and a text query $T$ as inputs, the studied task aims to output a video segment $[j, k]$ ($j$ and $k$ indicate the start and end clip indices respectively) that semantically matches the query description. Our work focuses on the weakly supervised setting of this task. Specifically, only a set of $V$-$T$ pairs are provided but the video segment annotation for each pair is not available. Inspired by the observation that humans usually locate interest events during a long video with a heuristic search strategy, we propose to formulate this task as a Markov Decision Process. A Boundary Adaptive Refinement (BAR) framework is thus designed: starting from an initial segment, the reinforcement learning technique is utilized to refine its temporal boundary progressively.
%To imitate human's thinking paradigm, Boundary Iterative Refinement (BIR) framework is proposed to address the task of weakly supervised temporal language grounding.
The overall architecture of the proposed BAR framework is depicted in Figure \ref{fig:framework}.
As illustrated, this modular framework employs a context-aware feature extractor to encode the environment state into cross-modal contextual concepts.
The cross-modal alignment evaluator is crafted to provide a tailor-designed reward and the termination signal for the iterative refinement process.
An adaptive action planner is designed to reason the direction and amplitude of the action from contextualized observation adaptively, instead of shifting a fixed amplitude every step \cite{he2019read}.
%The alignment information is also adopted to guide the adaptive action planner to determine the action scale.
The details of these modularized components will be described in the following sections.

\subsection{Context-aware Feature Extractor}
The context-aware feature extractor takes a video-query pair ($V$-$T$) from the external environment and encodes it into the context-aware cross-modal concepts.
Each word in query $T$ is firstly encoded using GloVe \cite{pennington2014glove} embeddings and then fed into the GRU \cite{chung2014empirical} to capture long-range dependencies. The  summarized query representation $\bm{\mathrm{E}}$ is obtained from the last hidden state of the GRU.
A pre-trained video feature extractor (C3D \cite{tran2015learning} or TSN \cite{wang2016temporal}) is used to extract the clip-level feature for each video clip.
A video segment is represented as a set of clip features, i.e., $\bm{\mathrm{F}}=\{\bm{\mathrm{F_{1}}};...;\bm{\mathrm{F_{i}}};...;\bm{\mathrm{F_{M}}}\} \in \mathbb{R}^{d_k \times M}$. $\bm{\mathrm{F_{i}}} \in \mathbb{R}^{d_k}$  denotes the clip-level feature for the video clip $V_i$ and $M$ is the number of clips in the corresponding video segment.
At each time step, the updated boundary divides the whole video into three parts: left segment, current segment and right segment. And we collect all clip-level features within the corresponding boundary into a set to obtain the three corresponding segment-level features.
Rather than directly taking the current segment's feature as independent inputs \cite{he2019read}, this extractor also leverages context-aware contextual cues derived from other segments in the video (i.e., left and right segment feature) for state encoding.
Furthermore, the extractor explicitly involves the normalized boundary location $\bm{\mathrm{L_{t-1}}}$ into the encoded features to provide some notion of relative position:
\begin{equation}
\bm{\mathrm{L_{t-1}}} = [\frac{l^s_{t-1}}{N},\frac{l^e_{t-1}}{N}]
\label{eq2}
\end{equation}
where $l^s_{t-1}$ and $l^e_{t-1}$ denote the start and end clip indices of the boundary, respectively. $t = \{1, \cdots, T_{max}\}$ and $T_{max}$ indicates the maximum iterations in the refinement process.

\subsection{Cross-modal Alignment Evaluator}
The cross-modal alignment evaluator is designed specially to address two critical issues in our RL-based approach.
On the one hand, this evaluator is crafted to assign a target-oriented reward to address the difficulty that the adaptive action planner can not directly obtain a reliable reward function without granular boundary annotations.
On the other hand, this alignment evaluator manages to determine an accurate stop signal to terminate the refinement process.
Given a video segment, the dimension of each clip feature is reduced to the same as the summarized query representation $\bm{\mathrm{E}}$ via a filter function $\theta$, which consists of a fully-connected layer followed by ReLU \cite{krizhevsky2012imagenet} and Dropout \cite{srivastava2014dropout} function.
$\bm{\mathrm{E}}$ is taken to create a temporal attention over all video clips, which manages to emphasize crucial video clips and weaken inessential parts.
Concretely, the scaled dot-product attention mechanism \cite{luong2015effective} is utilized to obtain attention weight $a_{i}$ and the segment attention feature $\bm{\mathrm{A}}$:
\begin{equation}
a_{i} = softmax (\frac{\bm{\mathrm{E}} \odot \theta(\bm{\mathrm{F_{i}}})}{\sqrt{k}}),\quad \bm{\mathrm{A}} = \sum_{i=1}^{M} a_{i} \theta(\bm{\mathrm{F_{i}}})
\label{eq3}
\end{equation}
where $\odot$ indicates the dot product operation between two vectors. $k$ is the dimension of $\bm{\mathrm{E}}$.
% and $M$ is the clip number of the input video segment.
 Then the segment attention feature and query representation are mapped to a joint embedding space to compute the alignment score $S$:
 \begin{equation}
S = \mathrm{L2Norm}(\bm{\mathrm{A}}) \odot \mathrm{L2Norm}(\bm{\mathrm{E}})
\label{eq4}
\end{equation}
The alignment score can be regarded as a reward estimate to provide reliable reward.
%Figure \ref{fig:reward} illustrates how BAR obtains the reward via the cross-modal alignment evaluator.
Specifically, the evaluator measures the alignment score of the consecutive segment-query pairs, and assigns the corresponding reward $r_t$:
%\begin{equation}
%\begin{split}
%r_t =
%\begin{cases}
%\zeta & \text{$S^c_{t} > S^c_{t-1}$}\\
%-\zeta & \text{$S^c_{t} \le S^c_{t-1}$}
%\end{cases},
%\end{split}
%\label{eq4}
%\end{equation}
\begin{equation}
r_t = \mathrm{sign}(S^c_{t} - S^c_{t-1})
\label{eq4}
\end{equation}
where $S^c_{t}$ denotes the alignment score of the current segment and sentence query at time step $t$.
This reward function returns +1 or -1. Basically, if the next boundary has a higher alignment score than the current one, the reward $r_t$ of the action $a_t$ moving from the current window to the next one is +1, and -1 otherwise. Such binary rewards reflect more clearly which action can drive the boundary towards the ground-truth and thus facilitate the agent's learning.

%We define three differential metrics to represent the change of alignment score that is caused by action $a_t$:
%\begin{equation}
%\begin{split}
%&D^{cg}_t = [S^c_{t} - S^g_{t}]- [S^c_{t-1} - S^g_{t-1}] = S^c_{t} - S^c_{t-1} \\
%&D^{cl}_t = [S^c_{t} - S^l_{t}] - [S^c_{t-1} - S^l_{t-1}] \\
%&D^{cr}_t = [S^c_{t} - S^r_{t}] - [S^c_{t-1} - S^r_{t-1}]
%\end{split}
%\label{eq5}
%\end{equation}
%$D^{cg}_t$\\

\subsection{Adaptive Action Planner}
The adaptive action planner is designed to infer action sequences to refine the temporal boundary.
To get a fixed-length visual representation, we utilize a mean pooling layer over feature set $\bm{\mathrm{F}}$ of the global, current, left and right segment, obtaining the pooling features $\bm{\mathrm{F^{g}}}, \bm{\mathrm{f^c_{t-1}}}, \bm{\mathrm{f^l_{t-1}}}, \bm{\mathrm{f^r_{t-1}}}$ respectively.
Then the cross-gated interaction method \cite{feng2018video} is further adopted to enhance the effects of the relevant segment-query pairs. Concretely, the current pooling feature $\bm{\mathrm{f^c_{t-1}}}$ is gated by query representation $\bm{\mathrm{E}}$, and meanwhile the gate of $\bm{\mathrm{E}}$ depends on $\bm{\mathrm{f^c_{t-1}}}$:
\begin{equation}
\tilde{\bm{\mathrm{f^c_{t-1}}}} = \sigma (\bm{\mathrm{W}^{s}}\bm{\mathrm{E}}) \odot \bm{\mathrm{f^c_{t-1}}}, \quad  \tilde{\bm{\mathrm{E}}} = \sigma (\bm{\mathrm{W}^{v}}\bm{\mathrm{f^c_{t-1}}}) \odot \bm{\mathrm{E}}
\label{eq1}
\end{equation}
where $\bm{\mathrm{W}^{s}}$ and $\bm{\mathrm{W}^{v}}$ are parameter matrices and $\sigma$ denotes the sigmoid function.
These cross-modal features are then concatenated and fed into two cascaded fully-connected layers
%all the clip features of the video segment to obtain the segment pooling feature. Then this network concatenates the cross-modal concepts and feeds them into two cascaded fully-connected layers
$\phi$ to get the state activation representation $s_t$:
\begin{equation}
s_t = \phi (\tilde{\bm{\mathrm{E}}}, \tilde{\bm{\mathrm{f^c_{t-1}}}}, \bm{\mathrm{f^g}}, \bm{\mathrm{f^l_{t-1}}}, \bm{\mathrm{f^r_{t-1}}}, \bm{\mathrm{L_{t-1}}}).
\label{eq1}
\end{equation}
%where $I^g, I^c_{t-1}, I^l_{t-1}, I^r_{t-1}$ are the global pooling feature, current pooling feature, left pooling feature and right pooling feature, respectively.
Such contextual features encourage the planner to perform a left-right tradeoff on the video contents and infer a more accurate action. $s_t$ is further fed into a GRU cell to enable the agent to incorporate the memory information about the video segments that have been explored.
Then the output state of the GRU is followed by two separate fully-connected layers (i.e., actor and critic) to respectively estimate a policy function $\pi(a_{t}|s_{t})$ and a value approximator $v^{\pi}(s_{t})$.
A primitive action $a_t \in \mathcal{A}$ is sampled from the policy function $\pi(a_{t}|s_{t})$ in the training procedure.
In our work, the action space $\mathcal{A}$ is composed of four primitive actions: shifting the start/end point backward/forward $N/\nu$ clips. $\nu$ is an amplitude factor that empirically sets as:
\begin{equation}
\nu = \lfloor 10 \times (1 + 2 \times \mathrm{tanh}(S^c_{t} - S^g)) \rfloor_+,
\label{eq2}
\end{equation}
where $\lfloor \rfloor_+ $ denotes the lower bound of a positive integer. $S^g$ and $S^c_{t}$ denotes the global and current alignment score estimated by the alignment evaluator.
$\mathrm{tanh}$ is used to constrain the action amplitude to fluctuate around $N/10$ (an empirical number used in ~\cite{he2019read}). $S^g$ plays as a baseline of the alignment degree to determine $\nu$: when $S^c_{t}$ is lower, $\nu$ becomes smaller and the agent markedly shifts the boundary; when $S^c_{t}$ becomes higher, $\nu$ is larger and the boundary is marginally refined.
This adaptive setting enables the agent to determine the action amplitude based on the current observation, which is also in line with human habits.

The state-value $v^{\pi}(s_{t})$ predicted by the critic is the value estimation of the current state. Under the assumption that the critic produces the exact values, the actor is trained based on an unbiased estimation of the gradient.

\subsection{Training}
Due to its efficiency, the advantage actor-critic (A2C) \cite{sutton2018reinforcement} algorithm is chosen to train our adaptive action planner.
Multiple instance learning algorithm with a combined ranking loss  $\mathcal{L}_{rank}$ is designed to train the cross-modal alignment evaluator and context-aware feature extractor.
The total loss in BAR is summarized as:
\begin{equation}
 \mathcal{L} = \mathcal{L}_{A2C} + \eta \mathcal{L}_{rank},
\label{eq5}
\end{equation}
where $\mathcal{L}_{A2C}$ denotes the loss function in the A2C algorithm. $ \eta$ is a trade-off factor between the two losses.

\noindent\textbf{A2C Loss.}
The adaptive action planner runs $T_{max}$ steps for adjustment during training. Given a trajectory in an episode $\Gamma = \langle s_t, \pi(\cdot | s_t), v^{\pi}(s_{t}), a_t, r_t \rangle$, the loss function of the actor $\mathcal{L}_{actor}$ is formulated as:
\begin{equation}
\small
 \mathcal{L}_{actor}= -\sum_{t=1}^{T_{max}} [A^{\pi}(s_t, a_t)\mathrm{log} \pi(a_t|s_t) + \alpha H(\pi(a_t|s_t))],
\end{equation}
where $A^{\pi}(s_t, a_t)$ denotes the advantage function and the entropy $H()$ of the policy is introduced into the objective for improving exploration.
$A^{\pi}(s_t, a_t) = Q^{\pi}(s_t, a_t) - v^{\pi}(s_t)$ measures whether or not and how much the action is better than the policy's default behaviour. Temporal-difference (TD) learning is adopted to estimate the Q-value function $Q^{\pi}(s_t, a_t)$ by $k$-step returns with function approximation:
\begin{equation}
Q^{\pi}(s_t,a_{t})= \sum_{l=0}^{k-1}\gamma^{l}r_{t+l} +\gamma^{k}v^{\pi}(s_{t+k})
\end{equation}
where $\gamma$ is a constant discount factor.
It is noted that the BAR does not suffer from sparse reward issue during training since the reward can be obtained at every step.  To optimize the critic, we minimize the mean squared error (MSE) loss  $\mathcal{L}_{critic}$ between the Q-value function and the estimated value \cite{mnih2016asynchronous}.
And the total A2C loss is a combination of the losses from the actor branch and the critic branch: $\mathcal{L}_{A2C} = \mathcal{L}_{actor} + \mathcal{L}_{critic}$.
%\begin{equation}
% \mathcal{L}_{A2C} = \mathcal{L}_{actor} + \mathcal{L}_{critic}
%\label{eq8}
%\end{equation}

\noindent\textbf{Ranking Loss.}
In general, the content discrepancy between the inter-videos is higher than that within the intra-video. Hence we resort to multiple instance learning algorithm and first leverage coarse-level semantic concepts from the inter-videos to optimize the framework.
Concretely, given the global video feature $F^g$ and its query representation $E$ , it is expected that the alignment score $S(F^g, E)$ (positive pair) is higher than the score $S({F^g}', E)$ / $S(F^g, E')$  (negative pairs) for any video ${F^g}'$ / query $\tilde{E}$ taken from other sample pairs. The inter-video ranking loss ~\cite{schroff2015facenet} is thus defined as:
\begin{equation}
\begin{split}
 \mathcal{L}_{inter} = \sum\limits_{E'} [\epsilon +S(F^g, E')-S(F^g,E)]_+ \\ + \sum\limits_{{F^g}'} [\epsilon +S({F^g}',E)-S(F^g,E)]_+ ,
\end{split}
\label{eq:9}
\end{equation}
where $[x]_+$ denotes a ramp function defined by $\mathrm{max}(0,x)$ and $\epsilon$ indicates a margin.
 $S(F^g, E)$ and $S^g$ are equivalent.
The positive and negative pairs are obtained from the same mini-batch.
%The training data is fully shuffled to select the batch samples, which can introduce a high degree of data diversity to facilitate effective training.

Inter-videos generally include substantially broad semantic abstractions that are hard to distinguish similar contents in a specific video. To this end, we design the intra-video ranking loss $\mathcal{L}_{intra}$ to capture more subtle concepts in the intra-video to further optimize the network.
Expressly, if the score of any one of left, current and right segment-query pairs surpasses the global one during the refinement process, we assume this pair should have higher alignment score than the other two pairs:
\begin{small}
\begin{equation}
\begin{split}
 \mathcal{L}_{intra}&=
{\mathcal{\psi}(S^c_{t} > S^g)}\times([\epsilon + S^l_{t} -S^c_{t}]_+ + [\epsilon +S^r_{t} -S^c_{t}]_+ )\\
&+{\mathcal{\psi}(S^l_{t} > S^g)}\times([\epsilon + S^c_{t} -S^l_{t}]_+ + [\epsilon +S^r_{t} -S^l_{t}]_+)\\
&+{\mathcal{\psi}(S^r_{t} > S^g)}\times([\epsilon + S^c_{t} -S^r_{t}]_+ + [\epsilon +S^l_{t} -S^r_{t}]_+),
\end{split}
\label{eq10}
\end{equation}
\end{small}
where $S^l_{t}$ and $S^r_{t}$ are the alignment scores of the left segment-query pair and the right segment-query pair at the time step $t$, respectively.
$\psi$() is a binary indicator function. If the inequality in parentheses holds, $\psi$() will output 1, otherwise 0.
Specifically, when the score of a segment-query pair, say $S^c_t$, surpasses $S_g$, the optimization target is to increase the gap between $S^c_t$ and the other two ($S^l_t$ and $S^r_t$ ) by increasing $S^c_t$ or decreasing $S^l_t$ and $S^r_t$ . Noted that by lowering $S^c_t$ below $S_g$ might be another option, but this usually becomes increasingly impractical with the progress of inter-video training. In addition, when there exist more than one segment-query pairs of score surpass $S_g$, the optimization target of $\mathcal{L}_{intra}$ will usually guide the alignment evaluator to suppress the score of the sub-optimal matching pair(s) to be lower than $S_g$ and at the same time drive the action planner to adjust the boundary.
Intuitively, $\mathcal{L}_{intra}$ encourages the text query to be closer to a semantically matched video moment than other possible moments from the same video, which contributes to obtaining a content-aware alignment score.

$\mathcal{L}_{intra}$ manages to i) widen the score gap between matched and unmatched segment-query pair to increase the confidence of the alignment evaluation; ii) improve the reward calculation by affecting the alignment evaluator to drive the action planner to achieve better temporal boundary adjustments. To sum up, the combined ranking loss $\mathcal{L}_{rank}$ is defined as:
\begin{equation}
 \mathcal{L}_{rank} =  \mathcal{L}_{inter} + \lambda \sum_{t=1}^{T_{max}} \mathcal{L}_{intra},
\label{eq11}
\end{equation}
where $\lambda$ is a weighting parameter to achieve a ranking loss  trade-off between the intra-video and the inter-video.
%As a weakly supervised setting, only the calculation of $\mathcal{L}_{inter}$ has a reliable video-level ground truth.
In the early stage of this collaborative training scheme, it is very unlikely that the score of a segment-query pair exceeds $S_g$ and $\mathcal{L}_{intra}$ tends to 0, hence $\mathcal{L}_{inter}$ plays a dominant role that learns to transfer the matching between video-query pair to segment-query pair. 
%Noted that this is achievable and is the core idea of weak supervision, imagining that there are multiple videos in the training set that have intersecting partial semantic descriptions. 
As the training progresses, $\mathcal{L}_{inter}$ converge gradually and it is more common for the score of segment-query pair to exceed $S_g$, $\mathcal{L}_{intra}$ begins to play a critical role.

\noindent\textbf{Alternating Update.}
BAR is trained from scratch and an alternating update strategy is applied to facilitate stable training.
Specifically, for each set of $2K$ iterations, we first fix the parameters of the action planner and employ $\mathcal{L}_{rank}$ for model optimization. This setting guarantees a trustworthy initial reward for the action planner.
When $K$ iterations are reached, we fix the parameters of the alignment evaluator and feature extractor, and switch $\mathcal{L}_{rank}$ to $\mathcal{L} {A2C}$ to optimize the action planner for $K$ more iterations. This alternating update mechanism repeats until the model converges. 

% The whole training algorithm is outlined in the Algorithm \ref{alg:1}.
% \begin{algorithm}[t]
% 	\caption{Alternating Update Training Procedure.}
% 	\begin{algorithmic}[1]
% 		\State Randomly initialize adaptive action planner parameters $\theta$.
% 		\State Randomly initialize alignment evaluator parameters $\omega$.
% 		\For {$iter=1,...I_{max}$} ($iter$: iterations)
% 		\State $iter= iter\bmod 2K$;
% 		\State Randomly sample a mini-batch.
% 		\If {$iter$ $<$ $K$}
% 		\State Freeze $\theta$ and Compute $S^g$.
% 		\For {$t=1,...,T_{max}$}
% 		\State Compute $S^c_{t}$ and $\nu$.
% 		\State Get $\hat{a}_{t}$ from $\pi$: $\hat{a_t}=\mathop{\arg\max}\pi(a_t|s_t)$.
% 		\EndFor
% 		\State Compute $\mathcal{L}_{inter}$, $\mathcal{L}_{intra}$ and $\mathcal{L}_{rank}$.
% 		\State Update $\omega$ by minimizing $\mathcal{L}_{rank}$.
% 		\ElsIf {$iter$ $\ge$ $K$}
% 		\State Freeze $\omega$ and compute $S^g$.
% 		\For {$t=1,...,T_{max}$}
% 		\State Compute $S^c_{t}$ and $\nu$.
% 		\State Sample $a_t$ from $\pi$: $a_t \sim \pi(a_t|s_t)$.
% 		\State Compute $S^c_{t}$, $S^c_{t-1}$ and get $r_t$.
% 		\EndFor
% 		\State Compute  $\mathcal{L}_{actor}$,  $\mathcal{L}_{critic}$ and $\mathcal{L}_{A2C}$.
% 		\State Update $\theta$ by minimizing $\mathcal{L}_{A2C}$.
% 		\EndIf
% 		\EndFor
% 	\end{algorithmic}
% 	\label{alg:1}
% \end{algorithm}

\begin{table*}[t]
\centering
\caption{The performance comparison (in \%) of the state-of-the-art methods in fully supervised and weakly supervised  setting. ``-'' indicates that the corresponding values are not available.}
\vspace{-5pt}
 \begin{tabular}{c|c|c|c|c|c|c|c}
\toprule
\multicolumn{3}{c|}{} & \multicolumn{3}{c|}{Charades-STA \cite{gao2017tall} }  & \multicolumn{2}{c}{ActivityNet \cite{krishna2017dense}}\\ \hline
Supervision & Feature &Baseline   &tIoU@0.7 & tIoU@0.5 & tIoU@0.3 & tIoU@0.5 & tIoU@0.3 \\ \hline
\multirow{12}{*}{Full Supervision} &\multirow{10}{*}{C3D} 
&ROLE \cite{liu2018cross}, ACM MM 2018 &-&12.12 &25.26 &-	&- \\
& &MCN \cite{anne2017localizing}, ICCV 2017 &4.44	&13.66 & 28.99& 10.17&22.07\\
& &CTRL \cite{gao2017tall}, ICCV 2017 &8.89	&23.63& -&14.36	&29.10\\
& &ACRN \cite{liu2018attentive}, SIGIR 2018 &9.65	&26.74 &47.64 &16.53	&31.75 \\
& &MAC \cite{ge2019mac}, WACV 2019 &12.23	&29.39& 53.34& - & - \\
& &SAP \cite{chen2019semantic}, AAAI 2019 &13.36	&27.42& - & -& -  \\
& &QSPN \cite{xu2019multilevel}, AAAI 2019 &15.80 & 35.60	&54.7  &27.70	&45.30   \\
& &ABLR \cite{yuan2018find}, AAAI 2019	&-&-&-&36.79	&55.67  \\
& &SM-RL \cite{wang2019language}, CVPR 2019 &11.17 &24.36 & - & -& -  \\
& &RWM \cite{he2019read}, AAAI 2019 &13.74	&34.12 &55.16 & 34.91 & 53.00  \\ \cline{2-8}
& TSN &RWM \cite{he2019read}, AAAI 2019 &17.72	&37.23 &61.73 & 37.46 & 57.29  \\ \cline{2-8}
& I3D &MAN \cite{zhang2019man}, CVPR 2019	&22.72&46.53&-&-&- \\ \hline \hline
\multirow{6}{*}{Weak Supervision} &\multirow{5}{*}{C3D} & TGA \cite{mithun2019weakly}, CVPR 2019 &8.84 &19.94&32.14&-&- \\
&& WS-DEC \cite{duan2018weakly}, NIPS 2018 &- &-&-& 23.34&41.98 \\
 && WSLLN \cite{gao2019wslln}, EMNLP 2019 &- &-&-&22.70&42.80 \\
 && SCN \cite{lin2019weakly}, AAAI 2020 &9.97 &23.58&42.96& 29.22&47.23 \\
&&BAR (our) &\textbf{12.23} &\textbf{27.04}&\textbf{44.97}&\textbf{30.73}&\textbf{49.03} \\ \cline{2-8}
& TSN &BAR (our) &\textbf{15.97} &\textbf{33.98}&\textbf{51.64}&\textbf{33.12}&\textbf{53.41} \\
\bottomrule
\end{tabular}
\label{table:result1}
 \end{table*}

\subsection{Inference}
At each time step, BAR executes an action $\hat{a_t}$ via greedy decoding algorithm to adaptively adjust the temporal boundary. And the cross-modal alignment evaluator computes a score $S^c_{t}$ to provide confidence for alignment degree and termination.
Empirically, the final grounding result corresponding to the query usually occupies a reasonable and appropriate video length.
Hence to penalize the video segment with abnormal lengths, we propose to update the confidence score with a Gaussian penalty function as follows:
\begin{equation}
P_{t} = \frac{l^e_{t}-l^s_{t}}{N}-\delta, \quad \hat{S^c_{t}} = S^c_{t} e^{-\frac{P_{t}^2}{\tau}}
\label{eq11}
\end{equation}
where $\delta$ denotes the penalty factor corresponds to abnormal lengths. $\tau$ is a modulating factor that as $\tau$ increases the effect of the penalty degree is likewise decreased. The segment with the max $\hat{S^c_{t}}$ during testing is regarded as the final grounding result.

\section{Experiments}

\subsection{Datasets and Evaluation Metrics}
\noindent\textbf{Datasets}. We conduct extensive experiments on two benchmark datasets:  Charades-STA \cite{gao2017tall} and ActivityNet \cite{krishna2017dense}.
Charades-STA is extended from the Charades dataset \cite{sigurdsson2016hollywood} with generated sentence-clip annotations, which comprises a series of sentence-clip pairs with 12,408 for training and 3,720 for testing.
The average length of each video in this dataset is 29.8 seconds and the described clips are 8 seconds long in average.
ActivityNet dataset \cite{krishna2017dense} is introduced to validate the robustness of the proposed model with longer and more diverse videos.
It contains 37,421 and 17,505 video-sentence pairs for training and testing. The average duration of the videos is 2 minutes and the described temporally annotated clips are 36 seconds long on average.
%We did not follow TGA \cite{mithun2019weakly} to perform the experiments on DiDeMo dataset \cite{anne2017localizing}.
%It is due to the fact that DiDeMo already provides 21 pre-defined segments for action localization, and the algorithm only needs to select the appropriate segment without optimizing the frame interval localization. It does not conform to the problem setting studied in this paper.

\noindent\textbf{Evaluation Metrics}. We adopt ``tIoU@ $\chi$'' to evaluate the grounding result.  ``tIoU @ $\chi$'' means the percentage of the queries that have temporal IoU larger than threshold $\chi$.

\subsection{Implementation Details}
We leverage C3D and the TSN model to encode video representation. The initial boundary is set to $L_{0} = [N/4; 3N/4]$.
$N/4$ and $3N/4$ denote the start and end clip indices of the boundary respectively.
%When the start point of the current segment reaches the beginning of the whole video, we set the feature of the first clip in the video clip set as the left segment feature. When the end point of the current segment reaches the end of the whole video, the last clip feature is treated as the right segment feature.
$T_{max}$ is set to 12 and the size of the hidden state in GRU is 1024. The batch size is 12 and the total loss is optimized via the Adam optimizer with the learning rate of 0.001. The margin $\epsilon$ in ranking loss is 0.2.
The hyper-parameters $\alpha$ and $\gamma$ is fixed to 0.1 and 0.4, receptively. The factor $\eta$ and $\lambda$ are empirically set to 1 and 0.1.
The modulating factor $\tau$ is set to 0.5 by cross validation. And penalty baseline factor $\delta$ is fixed to 0.35 and 1.0 receptively on Charades-STA and ActivityNet.
We use $K$ = 500 in the alternating update procedure.

\subsection{Comparison with the State-of-the-art}
We compare the proposed BAR with several state-of-the-art models based on the weakly-supervised and fully-supervised settings in Table \ref{table:result1}.
%We re-implement RWM \cite{he2019read} on both datasets, and TGA \cite{mithun2019weakly} on ActivityNet.
On the one hand, BAR significantly outperforms the weakly-supervised method and establishes new state-of-the-art performance on both datasets.
Employing the C3D based video feature, BAR boosts the tIoU@0.5 to 27.04\% and 30.73\%, with an improvement of 3.46\%, 1.51\% compared with SCN \cite{lin2019weakly} on the two datasets, receptively.
Furthermore, it manages to achieve 33.98\% (33.12\%) in tIoU@0.5 via more powerful TSN feature.
It reveals that our approach helps to better obtain accurate video segments.
%In the contrast experiment with TGA \cite{mithun2019weakly}, BAR can obtain a more substantial improvement on ActivityNet than Charades-STA.
%It may be due to the fact that this adaptive refinement is more prominent in the processing of longer video, while TGA suffers from exponentially sliding windows growth with respect to longer videos.
On the other hand, BAR even achieves better or comparable results than some fully-supervised methods.
For instance, BAR outperforms QSPN \cite{xu2019multilevel} by 3.03\% w.r.t tIoU@0.5 on the ActivityNet dataset.
This is an inspiring result as it reveals that our model can get impressive results via learning from massive coarse video-level annotations, which is of great benefit to practical application.

\subsection{Ablation Studies}
We perform extensive ablation studies and demonstrate the effectiveness of several essential components in BAR. The experiments are conducted on the Charades-STA with the TSN feature. The results are reported in Table \ref{table:result2}.

$\bullet$ \noindent \textbf{Effectiveness of Reinforcement Learning.}
More accurate measurement of the factual RL contribution is to directly remove it and use the generated proposals of an off-the-shelf weakly-supervised action localization method \cite{mithun2019weakly}. Hence we design a variant (abbreviated as ``Ours w/o RL'') to follow the above setting. We can observe that removing RL from BAR will lead to a noticeable drop in performance. For example, tIoU@0.5 declines from 33.98\% to 25.89\%. It reveals that the introduction of RL is fundamental and can bring more flexible and adaptable temporal proposals, this alone is an advantage that cannot be achieved with traditional two-stage frameworks, not to mention its high efficiency.

$\bullet$ \noindent \textbf{Effectiveness of Tailor-designed Reward.}
In order to validate that a target-oriented reward is essential for this task, we design a baseline (abbreviated as ``Ours w/ random reward'') that samples a random scalar value from the uniform distribution of [-1,1] as the reward for optimization. Table \ref{table:result2} shows that this baseline obtains an exceedingly inferior result, which is approximate to a stochastic one. It indicates that a tailor-designed reward is definitely necessary for the RL setting.

$\bullet$ \noindent \textbf{Effectiveness of Boundary Initialization.}
The initial boundary in this paper is fixed to $L_{0} = [N/4; 3N/4]$. To compare different boundary initializations, we design two baselines (denoted as ``Initial boundary [N/3; 2N/3]'' and ``Initial boundary [N/5; 4N/5]'') that sets the initial boundary as $[N/3; 2N/3]$ and $[N/5; 4N/5]$, respectively.
As reported in Table \ref{table:result2}, different boundary initialization is not sensitive to the performance of the algorithm, and all can obtain competitive experimental results, which reflects the robustness of BAR.

\begin{table}[t]
\centering
\caption{Performance of ablation models.}
\vspace{-3pt}
 \begin{tabular}{c|c|c|c}
\toprule
Metrics   & tIoU@0.7 & tIoU@0.5 & tIoU@0.3  \\ \hline
Ours w/o RL &12.37&	25.89&	45.36 \\
Ours w/ random reward  &5.76	&8.97	&28.82\\ \hline
Initial boundary [N/3; 2N/3] &15.72&	33.36&	51.33 \\
Initial boundary [N/5; 4N/5] &15.83&	33.47&	51.20 \\ \hline
Ours w/ N/5 amplitude &13.60&	31.88&	49.65 \\
Ours w/ N/10 amplitude &14.27&	32.02&	50.25 \\
Ours w/ N/15 amplitude &13.73&	31.66&	49.29 \\ \hline
Ours w/o context &13.62&	31.45&	49.22 \\
Ours w/o $\mathcal{L}_{intra}$ &14.24	&30.73	&46.82 \\
Ours w/ stop &10.13&	24.38&	43.22 \\
Ours w/o penalty &13.78	&30.97	&50.27 \\
Ours &\textbf{15.97} &\textbf{33.98}&\textbf{51.64} \\
\bottomrule
\end{tabular}
\label{table:result2}
 \end{table}

 $\bullet$ \noindent \textbf{Effectiveness of Adaptive Setting.}
Rather than shifting a fixed distance for each action, BAR can adaptively adjust its action amplitude according to the current state. To demonstrate the superiority of this adaptive setting, we design three variants (named as ``Ours w/ N/5 amplitude'', ``Ours w/ N/10 amplitude'' and ``Ours w/ N/15 amplitude'') that the agent shifts $N/5$, $N/10$ and $N/15$ clips at each step, respectively.
As summarized in Table \ref{table:result2}, ``Ours w/ N/10 amplitude'' when set with fixed adjustment strategy.
However, our approach with the adaptive setting manages to achieve more impressive performance, which reveals that this adaptive setting is more flexible and effective in our proposed framework.

%We also observe that ``Ours 3 initialization'' can achieve promising results, our model further promotes 0.46\% and 0.74\% on tIoU@0.7 and tIoU@0.5 respectively.
%It indicates that a suitable boundary initialization contributes to leveraging informative context concepts to achieve impressive grounding results.

\begin{figure}[t]
\centering
\subfigure[\footnotesize{Varying $\delta$ $\in$[0.1,0.5], $\tau$=0.5}]{
\includegraphics[width=0.47\linewidth]{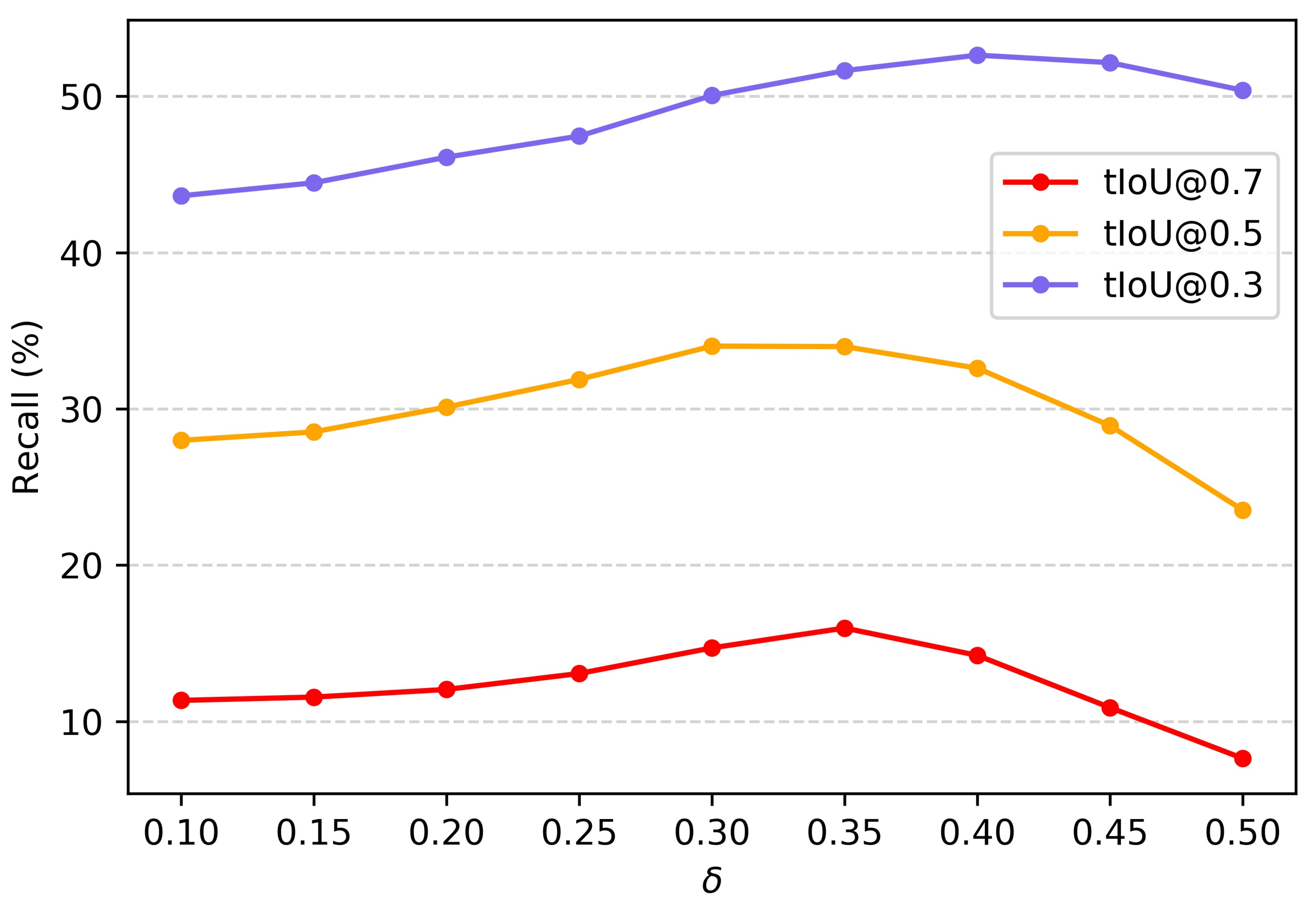}}
\subfigure[\footnotesize{Varying $\tau$ $\in$[0.1,5.0], $\delta$=0.35}]{
\includegraphics[width=0.47\linewidth]{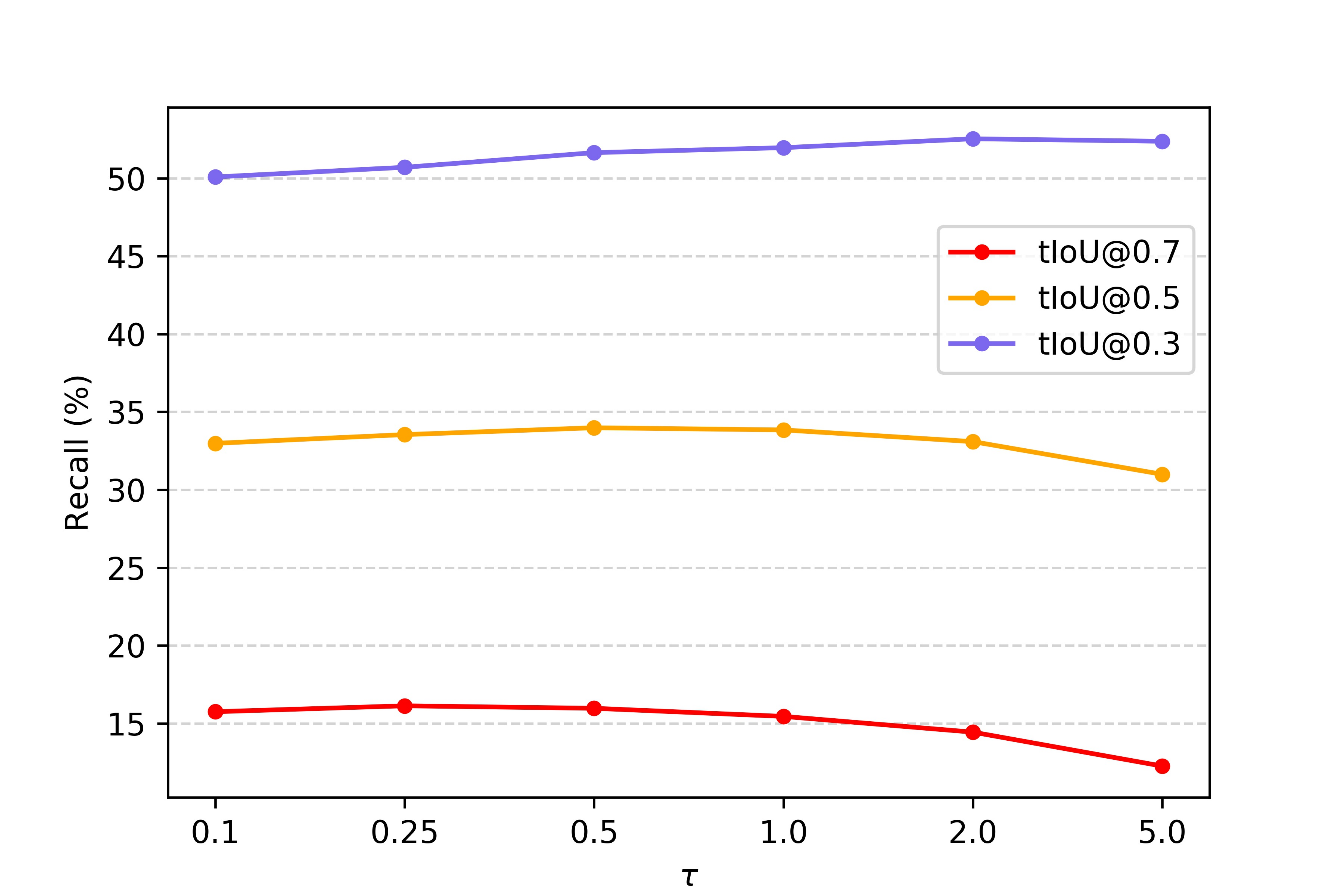}}
    \vspace{-3pt}
\caption{The performance curve of with varying hyper-parameters $\delta$ and $\tau$. Best viewed in color.}
\label{fig:penalty}
    \vspace{-15pt}
\end{figure}

\begin{figure*}
    \centering
    \includegraphics[width=0.83\linewidth]{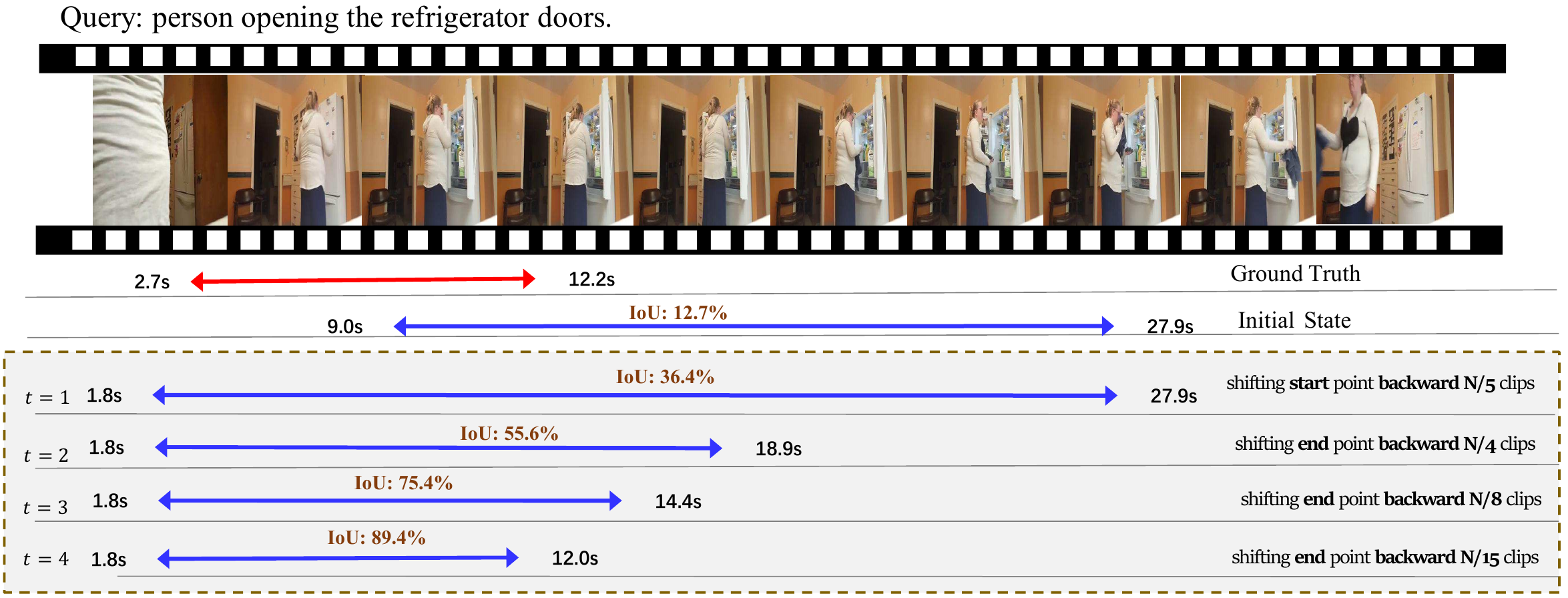}
    \vspace{-3pt}
    \caption{An illustration of how the proposed BAR framework accomplishes the task on Charades-STA.}
    \label{fig:example1}
    \vspace{0pt}
\end{figure*}

\begin{figure*}
    \centering
    \includegraphics[width=0.83\linewidth]{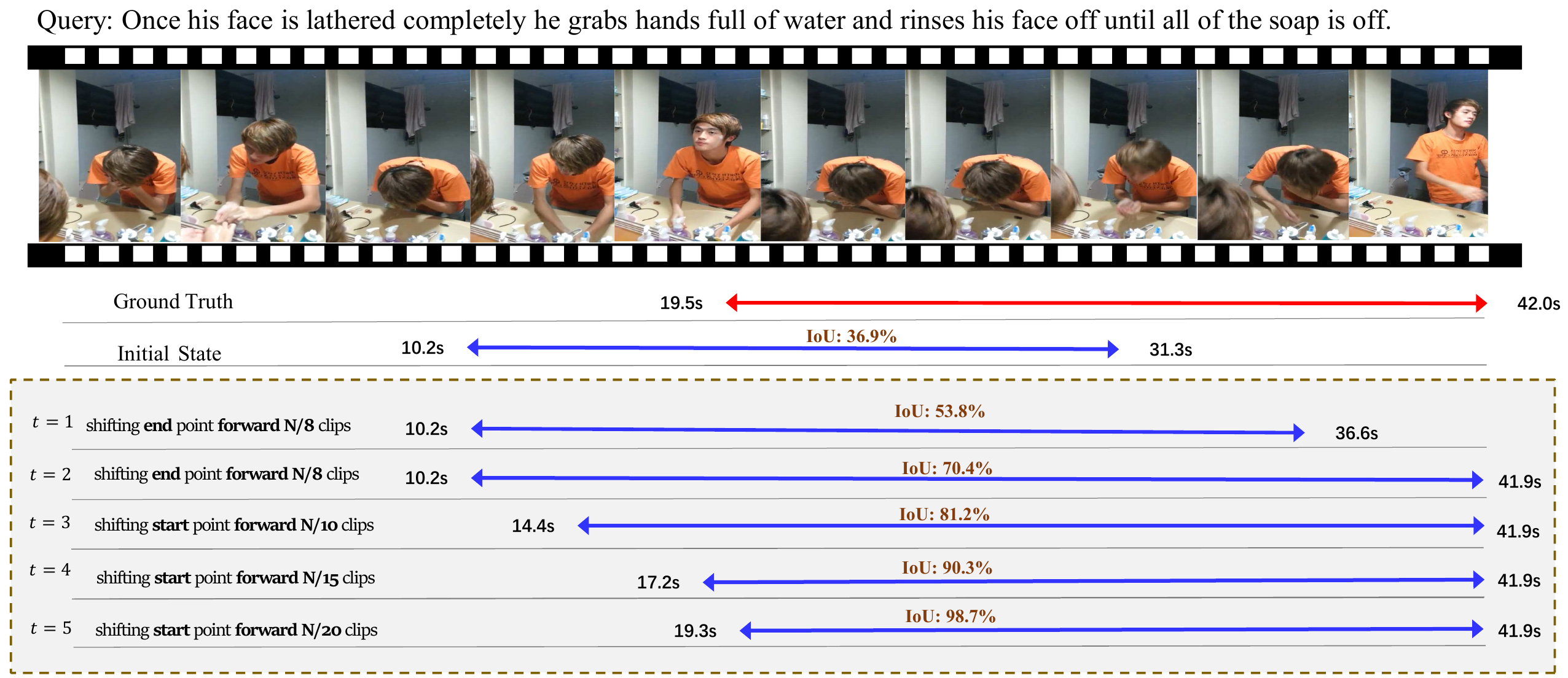}
    \vspace{-8pt}
    \caption{An illustration of how the proposed BAR framework accomplishes the task on ActivityNet.}
    \label{fig:example2}
   %\vspace{-8pt}
\end{figure*}

$\bullet$ \noindent \textbf{Effectiveness of Context Information.}
BAR additionally builds contextualized video representations for action decisions.
To investigate the effectiveness of the context information, we design a baseline that removes the context concepts ($f^l_{t-1}$, $f^r_{t-1}$) from $s_t$ in Equation \ref{eq1}, abbreviated as ``Ours w/o context''.
From Table \ref{table:result2}, we can see that although the model without context representation can still achieve promising results, our model with context involved gains 2.35\% and 2.53\% improvement w.r.t tIoU@0.7 and tIoU@0.5 respectively, which demonstrates that contextual concepts helps for obtaining more content-aware results.

%\noindent \textbf{Effectiveness of Attention Mechanism.}
%The dot-product attention mechanism is introduced to ensure the cross-modal alignment evaluator to focus more on informative video clips and alleviate the influence of noise on the alignment score.
%To evaluate the effectiveness of this attention mechanism, we design a baseline (denoted as ``Ours w/o attention'') that removes the attention mechanism in the alignment evaluator. Namely, the model ``Ours w/o attention'' directly maps the video pooling feature and query representation into the joint space to compute the alignment score.
%As reported in Table \ref{table:result2}, removing the attention mechanism results in dropping 1.60\% and 2.39\% on tIoU@0.7 and tIoU@0.5 respectively, indicating the effect of attention mechanism in our model in providing accurate alignment scores.

$\bullet$ \noindent \textbf{Effectiveness of Intra-video Ranking Loss.} 
%$\mathcal{L}_{intra}$ is creatively crafted to encourage the evaluator to capture fine-grained semantic concepts in a same video. 
To verify the effectiveness of the $\mathcal{L}_{intra}$, we construct a comparison variant that merely uses $\mathcal{L}_{inter}$ to optimize the evaluator, named as ``Ours w/o $\mathcal{L}_{intra}$''. Table \ref{table:result2} reveals that the grounding result suffers from an obvious drop without $\mathcal{L}_{intra}$. For example, tIoU@0.3 declines from 51.64\% to 46.82\%. Our approach with intra-video ranking loss manages to achieve more precise alignment scores and more accurate grounding results.
To further demonstrate the effectiveness of the alignment score $S$ obtained by our model, we additionally calculate the correlation coefficient (CC) between $S$ and ground-truth IoU. It shows that CC can reach 0.79, which reveals the obtained $S$ is reliable enough to correctly reflect the matching degree and infer the target-oriented rewards.

$\bullet$ \noindent \textbf{Analysis of Stop Signal.}
 We did not include an ending signal in the action space \cite{he2019read} as there is no absolutely reliable and stable internal segment-query matching that can help to effectively terminate the iteration. We further  introduce an alignment threshold as a stopping signal (abbreviated as``Ours w/ stop''), which led to inferior results.
In order to validate the significance of the length penalty strategy, we design a baseline that directly takes the score $S^c_{t}$ to determine the termination time, denoted as ``Ours w/o penalty''. The results indicate that this baseline suffers from performance degradation. It may be due to the fact that ``Ours w/o penalty'' tends to provide an excessive score when the length of the video segment is too long or too short.

Figure \ref{fig:penalty} depicts the performance curves with varying $\delta$ or $\tau$ respectively in the procedure of cross-validation. We can see that a factor $\delta$ with too large or too small value will lead to obvious performance decline, which reveals that a video with suitable length is more likely to produce impressive results.
A similar changing trend can be observed with varying $\tau$. It demonstrates that an appropriate gaussian penalty encourages the model to perform better.
we empirically observed that $\delta$=0.35 and $\tau$=0.5 contribute to obtaining the most promising performance in different levels of tIoU.
%$\bullet$ \noindent \textbf{Effectiveness of Alternating Update.}
%%Optimizing all three sub-networks together leads to unstable learning.
%To validate the necessity of the alternating update procedure, we design a baseline that optimizes the whole framework with total loss simultaneously, abbreviated as ``Ours w/o alternating''. Table \ref{table:result2} indicates that alternating update procedure can lead to a more impressive result. It may be due to the fact that the adaptive action planner in ``Ours w/o alternating'' fails to receive a reliable reward in the initial training phase.

\subsection{Efficiency}
To further investigate the efficiency of this boundary adaptive refinement process, we compare BAR with TGA \cite{mithun2019weakly} in terms of average running time and number of candidate segments. As summarized in Table \ref{table:result3}, BAR reduces the localization time and candidate boundaries by a sizeable margin.
Please notice that the boundary in BAR is equivalent to the temporal proposal number to some extent, but the ``boundary'' here is more flexible and adaptable.
%TGA \cite{mithun2019weakly} employs the sliding window based proposal-and-rank pattern and has to process extensive candidate segments one by one to localize queries.
BAR merely needs to refine an initial temporal boundary progressively, which manages to avoid redundant computations and employ a time-efficient and space-efficient manner.
Based on the above discussion, we can conclude that BAR is better than the previous competitive methods in both accuracy and efficiency.

\subsection{Qualitative Visualizations}
We illustrate two qualitative results in Figure \ref{fig:example1}, \ref{fig:example2} to show the whole process of how BAR obtains the described event location. We observe that our algorithm mainly performs optimization from coarse to fine. The agent will choose a larger movement adjustment at the initial stage of the iteration to quickly narrow the semantic difference between language and vision, and as the iteration progresses, the adjustment range of the movement will change rapidly to achieve local fine-tuning, this is also more consistent with humans performing cross-modal target retrieval.

 \begin{table}[t]
	\centering
	\caption{The average running time and number of candidate proposals to localize a moment in a video on Charades-STA.}
	\vspace{-5pt}
	\begin{tabular}{c|c|c}
		\toprule
		Methods  & Time(s) & Candidate Proposal Number \\ \hline
		TGA \cite{mithun2019weakly}  &0.104	&65.11 \\
		BAR (Ours) &\textbf{0.068}	&\textbf{1}	\\
		\bottomrule
	\end{tabular}
	\label{table:result3}
		
\end{table}

\section{Conclusions}
We propose a Boundary Adaptive Refinement framework that resorts to reinforcement learning to address the task of weakly-supervised temporal grounding of natural language in videos.
This refinement scheme completely abandons traditional sliding window-based solution patterns and contributes to obtaining more efficient, boundary-flexible and content-aware grounding results.
Extensive experiments show that our approach establishes new state-of-the-art performance on the widely used Charades-STA and ActivityNet datasets. Furthermore, our method even achieves a better result than some competitive fully-supervised methods.

\bibliographystyle{ACM-Reference-Format}
\balance
\bibliography{ACMMM_BAR}

%\appendix

\end{document}